\documentclass[pdflatex,sn-mathphys-num]{sn-jnl}

\usepackage{graphicx}%
\usepackage{multirow}%
\usepackage{amsmath,amssymb,amsfonts}%
\usepackage{amsthm}%
\usepackage{mathrsfs}%
\usepackage[title]{appendix}%
\usepackage{xcolor}%
\usepackage{textcomp}%
\usepackage{manyfoot}%
\usepackage{booktabs}%
\usepackage{algorithm}%
\usepackage{algorithmicx}%
\usepackage{algpseudocode}%
\usepackage{listings}%
\usepackage{float}%
\usepackage{placeins}%

\theoremstyle{thmstyleone}%
\theoremstyle{thmstyletwo}%

\theoremstyle{thmstylethree}%

\begin{document}

\title[Article Title]{Empowering Dysarthric Speech: Leveraging Advanced LLMs for Accurate Speech Correction and Multimodal Emotion Analysis}


\author*[1]{\fnm{Kaushal} \sur{Attaluri}}\email{iamkaushal49@gmail.com}

\author[2]{\fnm{Anirudh} \sur{Chebolu}}\email{ani1005sai@gmail.com}
\equalcont{These authors contributed equally to this work.}

\author[3]{\fnm{Sireesha} \sur{Chittepu}}\email{sireesha@staff.vce.ac.in}
\equalcont{These authors contributed equally to this work.}

\affil*[1]{\orgdiv{Software Engineer - Machine Learning and AI}, \orgname{Pegasystems Worldwide Pvt Ltd}, \city{Hyderabad},  \state{Telangana}, \country{India}}

\affil[2]{\orgdiv{Machine Learning  Engineer}, \orgname{Capital Quants Solutions}, \city{Hyderabad}, \state{Telangana}, \country{India}}

\affil[3]{\orgdiv{Asst Professor,Department of Information Technology}, \orgname{Vasavi College of Engineering}, \city{Hyderabad},  \state{Telangana}, \country{India}}


\abstract{Dysarthria or Dysarthric speech as called is kind of a motor speech disorder which is caused by neurological damage that affects the muscles for speech production, which results in slurred, slow, or difficult-to-understand speech which has been affecting millions of people worldwide including those with conditions such as stroke, traumatic brain injury, cerebral palsy, Parkinson's disease, and multiple sclerosis, dysarthric presents a significant communication barrier, which impacts the quality of life and social interaction of individuals. The entire aim of this paper is to come up with a mechanism that can recognize and translate the speech of dysarthric users and empower their ability to communicate effectively. In this paper, we are proposing a novel approach leveraging Advanced Large Language models for accurate speech correction and multimodal emotion analysis. Our methodology involves converting dysarthric speech to text using OpenAI's whisper model, followed by fine-tuned open source models such as LlaMa 3.1 70B and Mistral 8x7B models on Groq AI accelerators to predict the intended sentences from distorted input speech accurately. The entire dataset used was made by combining TORGO dataset with Google speech data and then manually labeling emotional contexts. Our framework identifies emotions such as happiness, sadness, neutral, surprise, anger and fear highlighting the potential understanding of dysarthric speech. Our approach effectively reconsturcts the intended sentences and detects emotions with high accuracy.}




\maketitle

\section{Introduction}\label{sec1}
Dysarthria is a motor speech disorder resulting from neurological damage that affects the muscles involved in speech production, such as the tongue, lips, vocal cords, and diaphragm. This condition often leads to slurred, slow, or difficult-to-understand speech, creating significant communication barriers for affected individuals. Dysarthria is not a singular condition but a symptom associated with various neurological disorders, including stroke, traumatic brain injury, cerebral palsy, Parkinson's disease, and multiple sclerosis. As a global issue, dysarthria affects millions of people worldwide, limiting their ability to communicate effectively and impacting their social interactions, quality of life, and overall well-being.

Existing methods to assist individuals with dysarthria in communication are often limited to conventional speech therapy or assistive communication devices, which may not be effective for all cases, especially where speech clarity is severely compromised. The challenge lies in creating solutions that not only improve speech intelligibility but also capture the underlying emotions conveyed by the speaker. Emotions play a crucial role in effective communication, and understanding them is vital for ensuring meaningful interactions. Addressing both aspects—speech correction and emotion recognition—can significantly enhance the overall communicative experience for dysarthric users.

Advancements in artificial intelligence (AI) and natural language processing (NLP) have opened new avenues for developing more sophisticated solutions to address dysarthria-related communication barriers. In recent years, Large Language Models (LLMs) such as OpenAI's Whisper, LLaMA 3.1, and Mistral 8x7B have demonstrated remarkable capabilities in understanding, generating, and interpreting human language. These models, especially when fine-tuned for specific tasks, offer promising potential to decode and reconstruct dysarthric speech accurately. Additionally, the deployment of these models on highly efficient hardware like GROQ LPUs further enhances their performance and applicability in real-time scenarios.

In this study, we propose a novel framework for dysarthric speech recognition and correction that leverages advanced LLMs for accurate speech-to-text conversion and sentence prediction. We use OpenAI's Whisper model to transcribe dysarthric speech into text, followed by fine-tuning LLaMA 3.1 (70B) and Mistral 8x7B models to predict the intended sentence from the distorted input. To address the multimodal aspect of communication, our approach also includes emotion recognition using the same LLMs to identify six primary emotions—sadness, happiness, surprise, anger, neutrality, and fear—present in the user's speech. We created a comprehensive dataset by integrating the TORGO dataset with Google Speech data and manually annotating emotional contexts to train our models for both tasks.

Our proposed solution demonstrates significant improvements in reconstructing intended speech and recognizing underlying emotions, achieving high accuracy levels when compared to actual speech data. By combining speech correction with emotion detection, we aim to provide a more inclusive and effective communication aid for dysarthric individuals. This research contributes to the field of speech pathology and artificial intelligence by offering a scalable, AI-driven approach to empower dysarthric users, ultimately enhancing their social and emotional well-being.

\section{Literature Survey}\label{sec2}

The paper \cite{efstathiadis2024llm}, investigates the use of large language models (LLMs) to improve speaker diarization accuracy. The authors propose a method that fine-tunes an LLM on a large dataset of transcribed conversations to score speaker diarization hypotheses. They evaluate their method on a benchmark dataset and report that fine-tuned LLMs can improve accuracy, but their performance is constrained to transcripts produced using the same ASR tool. To address this constraint, they develop an ensemble model that combines weights from three separate models, each fine-tuned using transcripts from a different ASR tool. The ensemble model demonstrates better overall performance than each of the ASR-specific models, suggesting that a generalizable and ASR-agnostic approach may be achievable. \\

The use of Large Language Models (LLMs) for emotion recognition from audio has been less explored compared to their use for text-based emotion recognition. This article \cite{wu2024beyond} investigates the effectiveness of LLMs for audio-based emotion recognition by using audio features as input. The authors find that LLMs can achieve good performance on emotion recognition when given audio descriptions, but the quality of the audio data significantly impacts the results. This research contributes to the growing field of audio-based emotion recognition using LLMs and highlights the importance of high-quality audio data for accurate emotion detection.  \\

ClozeGER \cite{hu2024listen}is a novel approach to ASR generative error correction that utilizes a multimodal LLM, SpeechGPT, to improve correction fidelity. By reformulating GER as a cloze test with logits calibration, ClozeGER addresses the limitations of traditional methods. This approach leverages the strengths of multimodal LLMs and provides clearer instructions to the model, resulting in more accurate and contextually relevant corrections. Experimental evidence demonstrates the effectiveness of ClozeGER on various ASR datasets, surpassing previous state-of-the-art methods.  \\

\textit{Multi-Stage LLM-Based ASR Correction} presents a novel approach \cite{pu2023multi} to address the over-correction issue in LLM-based ASR correction. By incorporating an uncertainty estimation stage and formulating the task as a multi-step rule-based LLM reasoning process, the proposed method achieves state-of-the-art performance, even in zero-shot settings. This work highlights the potential of rule-based LLM reasoning for future ASR correction tasks, while also emphasizing the need for further research to enable this approach in smaller and less capable LLMs.  \\

\textit{Multi-Stage LLM-Based ASR Correction} presents a novel approach \cite{li2024investigating}to address the over-correction issue in LLM-based ASR correction. By incorporating an uncertainty estimation stage and formulating the task as a multi-step rule-based LLM reasoning process, the proposed method achieves state-of-the-art performance, even in zero-shot settings. This work highlights the potential of rule-based LLM reasoning for future ASR correction tasks, while also emphasizing the need for further research to enable this approach in smaller and less capable LLMs. \\

In this work, \cite{ma2024correction} they propose a correction-focused LM training strategy to better boost performance on ASR fallible words. Correction-focused training relies on word-level ASR fallibility scores, which we define as the likelihood of being mis-recognized by a given ASR. We leverage LLMs to assist the data preparation for correction-focused training. The LLMs are fine-tuned to serve as fallibility score predictors and text generators simultaneously through multi-task fine-tuning, which significantly improves data preparation efficiency. Experiment results show the profound effectiveness of correction-focused training, indicating the informativeness of fallibility scores. In the future, we will continue to explore this direction by leveraging more information from ASR decoding results, such as confusion word pairs. \\

In this survey, we provide a comprehensive overview of recent advancements in Multimodal Large Language Models (MM-LLMs). We delve into their architecture, training pipelines, and state-of-the-art models, highlighting their unique formulations and performance on various benchmarks. By offering insights into the capabilities and future directions of MM-LLMs, we aim to contribute to the ongoing progress in this rapidly evolving field.

\section{Methodology}\label{sec3}

\subsection{Data Pre-processing}\label{subsec2}
The data information was taken from the Torgo dataset. The torgo dataset consists of only singular dysarthric word, where the frequency of the word is not consistent and using the available Torgo dataset, we have created our own dataset using the available torgo dataset. We had multiple steps in the creation of the dataset which is discussed in this paper step by step in the below subsections.

\subsubsection{Step-1: Using GPT-4 to convert to sentences}
So, as discussed initially only the word was available in the torgo dataset which which was dysarthric where the voice was could not be detected. So we used GPT-4 API services and have written a prompt to get them converted to understandable english sentences.
We used GPT-4 for this because it gave us the best results for text generation. \\

\textbf{For example:} In the TORGO dataset, the irregular word is, **"cash"**. \\

\textbf{After GPT-4.o text generation:} The sentence is formed as:  \textbf{\textit{*I always keep some **cash** in case of emergencies.*}} \\

So here the word cash is replaced by a space which indicates, that it is not understood by the listener.

\subsubsection{Step-2: Using Google Speech to convert them to audio files}
Later after getting the overall sentences in text and the existing irregularity in the word, we used Google speech to convert these sentences to speech which will act as the input to our model.

\subsection{Speech To Text}\label{subsec2}

\begin{figure}[h]
    \centering
    \includegraphics[width=1.0\textwidth]{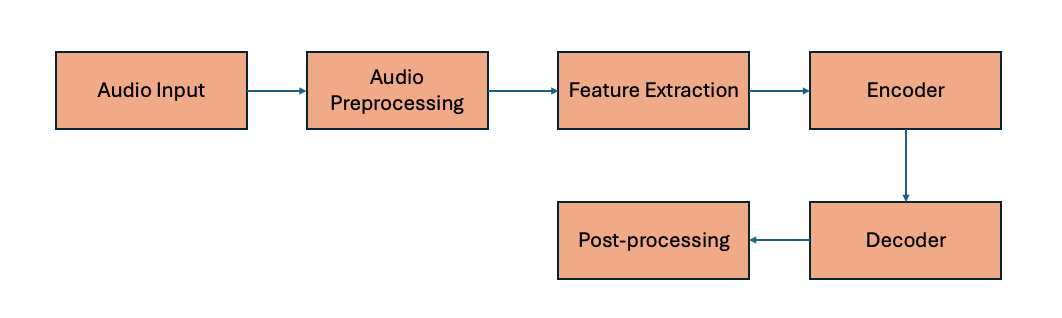}
    \caption{\centering{Architecture Flow of Whisper}}
    \label{fig: Architecture Flow}
\end{figure}
After getting the input format to speech and creating the dataset, we had to convert the existing speech format to text, for this conversion, we are using OpenAI's whisper which is an automatic speech recognition system. The great advantage of whisper is, it can convert speech to text in multiple languages, It is multilingual and it is open source. \\

\subsection{GPT 4.o}\label{subsec3}
\begin{figure}[h]
    \centering
    \includegraphics[width=1.0\textwidth]{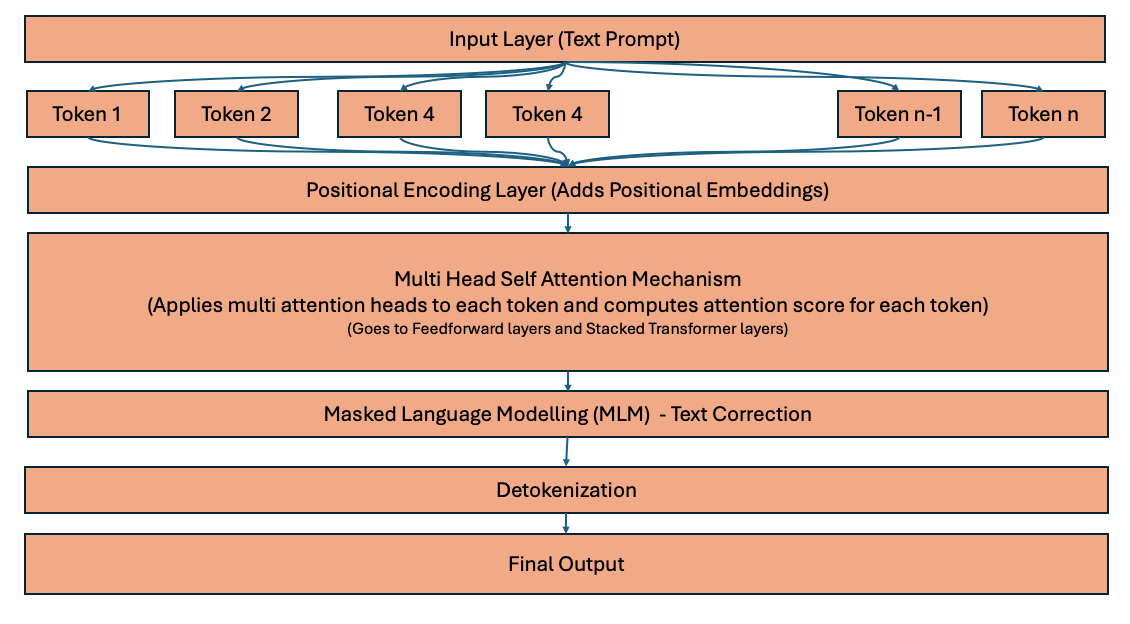}
    \caption{\centering{GPT-4.o Architecture}}
    \label{fig: Architecture Flow}
\end{figure}
After converting the speech to text, we used the GPT 4.o model to get the predicted sentence. GPT 4.o is the part of the (Generative Pre-trained Transformer) series which was developed by OpenAI building upon the high success of it's succcessors.
GPT-4.o was launched in May 2024 with over 200 billion parameters and is considered to be the State of the Art (SOTA) model. It has very good understanding of nuanced queries. \\

GPT-4.o's Architecture is built on a transformer model, which uses self-attention mechanisms to understand the relationship between words which is crucial for interpreting dysarthric speech, where missing or distorted words creates created discontinuities.

The key aspects of GPT-4.o Architecture are:

\begin{itemize}
    \item \textbf{Self-Attention Mechanisms}: The ability to weigh the importance of each word in the input allows GPT-4.o to handle incomplete or distorted speech. In cases where dysarthric speech omitted words or phrases, GPT-4.o was able to infer the likely missing elements by considering the context of the surrounding words, which is essential when processing challenging speech patterns.

    \item \textbf{Context Length}: GPT-4.o’s increased capacity to process longer text sequences proved beneficial in handling dysarthric speech inputs. Many sentences required GPT-4.o to maintain coherence over longer sequences to correctly interpret and reconstruct the speech. Its longer context window enabled the model to keep track of complex sentence structures, even when parts of the speech were unclear or missing.

    \item \textbf{Pre-training on Large, Diverse Datasets}: GPT-4.o’s pre-training on a vast array of text data from diverse domains made it more capable of understanding less common speech patterns, such as those produced by dysarthric individuals. This broad pre-training helped the model adapt to the unique linguistic challenges presented in our use case, improving its predictive accuracy.

    \item \textbf{Multimodal Learning Capabilities}: Although we used GPT-4.o primarily for text-based tasks, its multimodal capabilities ensure that it can handle inputs from different formats, such as audio and text. This inherent flexibility allowed it to integrate seamlessly with our speech-to-text pipeline, enhancing its ability to interpret dysarthric speech effectively.
\end{itemize}

Overall GPT-4.o gave us good accuracy when compared to other Large Language Models when integrated into our workflow for predicting the correct sentence and recognizing the emotion behind the sentence delivering a high inclusive communication solution for dysarthric users.

\subsection{Llama 3.1-70b}\label{subsec6}
The large language model used for understanding the dysarthric word, we used Llama 3.1-70b-Versatile model developed by Meta AI which was launched in July 2024 which already has over 2 million downloads on hugging face, the reason we used this is that as of now it is one of the best open-source models and is ranked among the top base models in the Open LLM leaderboard on huggingface.
We used this model as it is one of the best LLMs right now especially for text generation use case and when we tried samples, it gave us the best results.
\begin{itemize}
    \item \textbf{Self-Attention Mechanisms}: LLaMA 3.1-70B utilizes an advanced self-attention mechanism, which is key for interpreting incomplete or distorted speech. This ability to focus on the contextual importance of words allowed the model to predict missing elements in dysarthric speech, enhancing the clarity of sentence reconstructions.

    \item \textbf{Fine-Tuning for Dysarthric Speech}: Fine-tuning the model on our customized dataset, based on the TORGO dataset and Google Speech, was crucial. This process allowed LLaMA 3.1-70B to learn the specific linguistic patterns present in dysarthric speech, such as slurred words, omissions, and irregular phrasing. Fine-tuning helped align the model’s predictions with the complexities of our task, boosting its accuracy to 74.6\%. This approach made the model more specialized in handling real-world cases of dysarthric speech, leading to more accurate reconstructions.

    \item \textbf{Long-Range Dependencies}: The ability to process long-range dependencies allows the model to retain context over extended sequences. Dysarthric speech often includes incomplete phrases, and LLaMA 3.1-70B’s ability to consider the entire sentence context helped the model fill in gaps, improving overall sentence coherence.

    \item \textbf{Improvements Over LLaMA 2 and 3}: LLaMA 3.1 introduces several improvements over its predecessors:
    \begin{itemize}
        \item \textit{Increased Parameter Efficiency}: Compared to LLaMA 2 and 3, version 3.1 has better parameter optimization, reducing computational overhead while maintaining or exceeding performance.
        \item \textit{Enhanced Pre-Training Data}: LLaMA 3.1 benefits from pre-training on a broader and more diverse dataset than previous versions, which enhanced its ability to generalize, especially when handling non-standard or incomplete speech inputs.
        \item \textit{Improved Fine-Tuning Mechanisms}: The fine-tuning architecture of LLaMA 3.1 is more flexible, allowing for greater adaptability to specific tasks like dysarthric speech recognition. This flexibility helped us leverage domain-specific data effectively, leading to higher accuracy compared to using LLaMA 2 or 3.
    \end{itemize}

    \item \textbf{Performance in Our Use Case}: Through these architectural improvements and fine-tuning, LLaMA 3.1-70B was able to accurately predict and reconstruct sentences from dysarthric speech with 74.6\% accuracy. It also proved valuable in handling longer speech sequences and correctly inferring missing or distorted words, making it a key component in our speech-to-text pipeline.
\end{itemize}

LLaMA 3.1-70B's advances in efficiency, enhanced training data, and adaptability, combined with our fine-tuning process, allowed the model to outperform its predecessors in handling the complexities of dysarthric speech.

\subsection{Mixtral-8x7b-32768}\label{subsec7}
Mistral-8x7b-32768 is a cutting-edge large language model that excels in high-performance language understanding tasks. This model was key in achieving a 70.2\% accuracy in our dysarthric speech recognition project. Here, we discuss its history, architectural innovations, and how it was fine-tuned for our use case.

\begin{itemize}
    \item \textbf{Model History}: Mistral, a newer generation of large language models, has been developed with a focus on scaling up model size while improving computational efficiency. The Mistral-8x7b model with 32,768 tokens is designed for long-context tasks, enabling robust understanding of complex sequences, which makes it highly relevant for real-world speech processing applications like dysarthric speech recognition.

    \item \textbf{Architecture and Technical Innovations}:
    \begin{itemize}
        \item \textit{Hybrid Transformer Mechanism}: The Mistral-8x7b-32768 employs a hybrid transformer architecture with multi-head self-attention and feedforward layers that efficiently handle extremely large token inputs (up to 32,768 tokens). This architecture allows the model to process extended input sequences from dysarthric speech, where key words might be missing or garbled, helping it predict the correct context and reconstruct incomplete sentences.
        \item \textit{Sparse Attention and Memory Efficiency}: To handle longer input sequences efficiently, Mistral-8x7b leverages sparse attention mechanisms. These mechanisms ensure that the model selectively attends to the most relevant parts of the input, reducing computational load while retaining context over large inputs, which is critical for handling the irregularities of dysarthric speech.
        \item \textit{Parallelized Training on Groq LPU}: We fine-tuned the Mistral-8x7b-32768 model using the Groq LPU (Language Processing Unit), which allowed us to parallelize computations effectively, resulting in faster convergence during fine-tuning. Groq's parallel architecture is particularly beneficial for handling models like Mistral, as it enables real-time speech processing by accelerating inference and improving overall latency in model predictions.
    \end{itemize}

    \item \textbf{Fine-Tuning and Application in Dysarthric Speech Recognition}: 
    The fine-tuning process involved adapting Mistral-8x7b to the specific speech irregularities found in the dysarthric speech dataset we created from TORGO and Google Speech. This allowed the model to learn patterns such as slurring, missing words, and distortions, significantly improving its accuracy in predicting and reconstructing sentences. The model's ability to handle long input sequences proved valuable, as dysarthric speech often required context retention over longer segments of text. Our fine-tuning on Groq helped optimize its performance for real-time use, achieving a 70.2\% accuracy in sentence reconstruction.

    \item \textbf{Performance in Our Use Case}: By leveraging Mistral-8x7b-32768's architecture and advanced attention mechanisms, we were able to process incomplete and distorted speech inputs from dysarthric individuals, reconstructing sentences with a reasonable level of accuracy. The model's combination of sparse attention, memory efficiency, and large token capacity made it an ideal fit for handling the intricacies of dysarthric speech in our pipeline.
\end{itemize}

Mistral-8x7b-32768’s technical capabilities, combined with the fine-tuning on Groq LPU, made it a critical part of our dysarthric speech recognition solution, achieving a 70.2\% accuracy rate in handling complex speech patterns.

\subsection{Using QLoRA to Fine-Tune LLMs}\label{subsec8}
Fine-tuning large language models (LLMs) such as LLaMA 3.1 and Mistral-32B for domain-specific tasks is critical for improving performance in specialized tasks like dysarthric speech recognition. We employed QLoRA (Quantized Low-Rank Adaptation), a technique that enables efficient fine-tuning by adapting smaller models to larger LLMs. This section explains the QLoRA approach and demonstrates how it was used to fine-tune LLaMA 3.1 and Mistral-32B on the TORGO dataset for improving both speech and emotion detection accuracy.

\subsubsection{LoRA and QLoRA: An Overview}
\textbf{LoRA (Low-Rank Adaptation)} \cite{lora2021} is a method for efficient model fine-tuning. Instead of updating all model weights during fine-tuning, LoRA inserts low-rank decomposition matrices into the weights of transformer layers. These matrices are smaller, which allows the model to adapt to new tasks without retraining all the parameters. Given a weight matrix $W \in \mathbb{R}^{d \times k}$, LoRA approximates $W$ as:
\[
W' = W + BA
\]
where $A \in \mathbb{R}^{r \times k}$ and $B \in \mathbb{R}^{d \times r}$ are low-rank matrices (with rank $r \ll d$).

\textbf{QLoRA (Quantized LoRA)} builds upon LoRA by applying 4-bit quantization to the model weights before fine-tuning. This significantly reduces memory usage, enabling fine-tuning of extremely large models on commodity hardware. In QLoRA, the forward pass uses quantized weights, while LoRA layers allow gradients to flow during backpropagation.

\subsubsection{Architecture of QLoRA}
\begin{figure}[h!]
    \centering
    \includegraphics[width=0.8\textwidth]{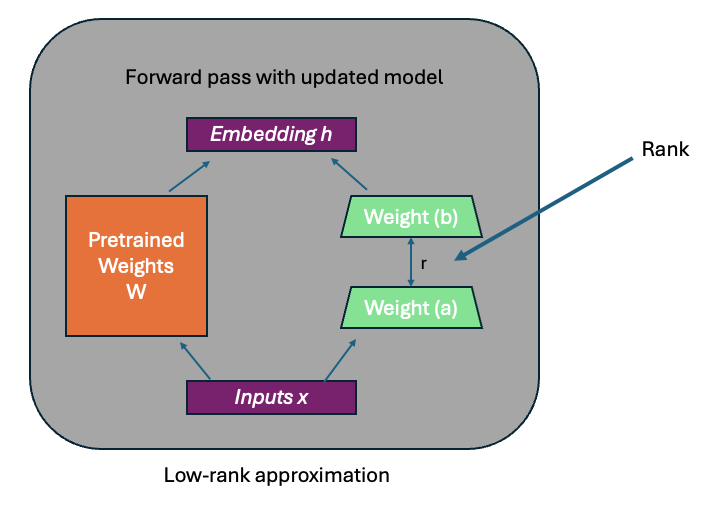}
    \caption{{QLoRA Architecture}}
    \label{fig:qlora_architecture}
\end{figure}

The architecture of QLoRA involves quantizing the LLM's model weights to reduce memory requirements and applying low-rank adaptation to specific layers. This is particularly important in transformer architectures, where each transformer block consists of multi-head attention and feed-forward layers. In QLoRA:
\begin{itemize}
    \item The full model weights are quantized to 4-bit integers.
    \item Low-rank matrices are added in specific layers to capture task-specific knowledge.
    \item Only the LoRA parameters are updated during training, while the quantized weights remain fixed.
\end{itemize}

The core architecture of QLoRA can be depicted as:
\[
W_q' = W_q + BA
\]
where $W_q$ are the quantized model weights, and $BA$ are the low-rank updates, as in the original LoRA architecture.

\subsubsection{Fine-Tuning LLMs using QLoRA}
Fine-tuning an LLM using QLoRA involves three key steps:
\begin{enumerate}
    \item \textbf{Model Preparation:} The weights of the LLM (e.g., LLaMA 3.1, Mistral-32B) are quantized to 4-bit precision. This drastically reduces the memory footprint.
    \item \textbf{Low-Rank Adaptation:} LoRA layers are inserted into the transformer blocks of the model. These layers are initialized with small, low-rank matrices.
    \item \textbf{Training on Task-Specific Data:} During training on the TORGO dataset, only the LoRA parameters are updated. The quantized weights remain unchanged, ensuring memory efficiency.
\end{enumerate}

The key advantage of this approach is that the effective model size remains manageable, allowing fine-tuning on a single GPU while retaining high performance.

\subsubsection{QLoRA for Dysarthric Speech Recognition}

By employing QLoRA, we achieved significant improvements in both speech and emotion recognition for dysarthric speech using the TORGO dataset. Quantized fine-tuning enabled better generalization without overfitting, and the low-rank adaptation matrices captured the variability in dysarthric speech patterns. The overall accuracy improvements for models fine-tuned with QLoRA were as follows:
\begin{itemize}
    \item \textbf{Speech Detection:} For LLaMA 3.1, we observed a boost from 68.2\% to 74.6\% after QLoRA fine-tuning.
    \item \textbf{Emotion Detection:} The model's ability to classify emotions improved from 81\% to 89\% for fine-tuned models on dysarthric data.
\end{itemize}

\subsubsection{Formula for Fine-Tuning}
To quantify the improvement from QLoRA, let $Acc_{baseline}$ represent the baseline accuracy of the pre-trained model, and $Acc_{fine-tuned}$ be the accuracy after fine-tuning with QLoRA. The improvement can be calculated as:
\[
\Delta Acc = Acc_{fine-tuned} - Acc_{baseline}
\]
For example, for LLaMA 3.1, $\Delta Acc$ for speech recognition is:
\[
\Delta Acc = 74.6\% - 68.2\% = 6.4\%
\]
This illustrates the effectiveness of QLoRA in boosting model performance while maintaining resource efficiency.

\begin{figure}[h!]
    \centering
    \includegraphics[width=0.8\textwidth]{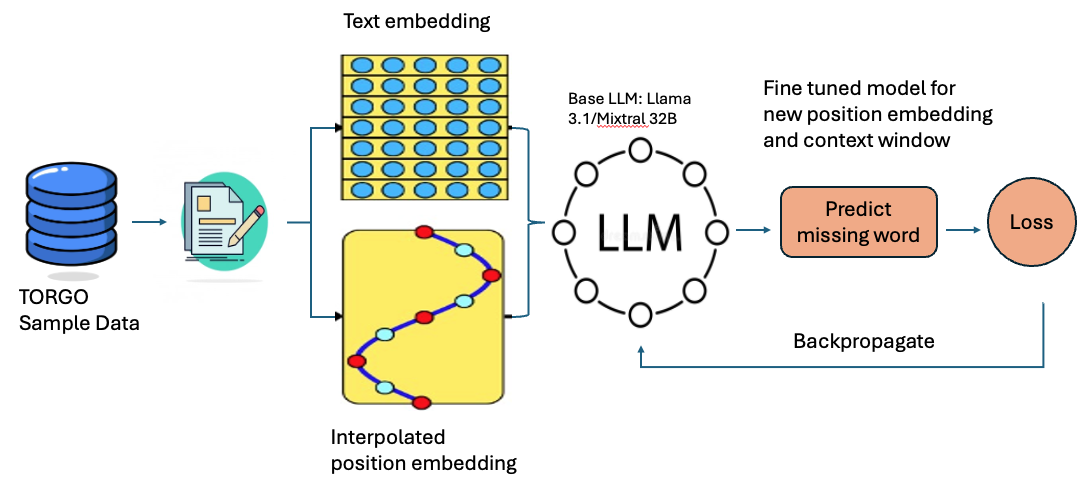}
    \caption\centering{{Fine-Tuning LLM Architecture Flow}}
    \label{fig:finetuning_llms}
\end{figure}

\begin{equation}
W_q' = W_q + BA
\end{equation}

\subsection{Emotion Recognition}\label{subsec9}

In addition to sentence reconstruction, our system performs emotion recognition by classifying the emotional state of dysarthric speakers into six classes: \textit{neutral}, \textit{happy}, \textit{sad}, \textit{anger}, \textit{surprise}, and \textit{fear}. Accurately detecting the emotion behind dysarthric speech is essential, as it provides contextual understanding that goes beyond syntactic correctness, enhancing interaction between dysarthric individuals and assistive technologies.

\textbf{Emotion Recognition Pipeline}: For the emotion recognition task, we manually labeled the emotional context of each sentence in our dataset. After sentence reconstruction, the predicted sentences were passed through three large language models (GPT-4.0, LLaMA 3.1-70B, and Mistral-8x7b-32768) to classify the emotional state of the speaker.

\textbf{Technical Approach to Emotion Classification}: All three models leverage transformer-based architectures that make them adept at handling language modeling and emotion recognition. Here’s a breakdown of how the transformer architecture enables this functionality:

\begin{itemize}
    \item \textbf{Self-Attention Mechanism for Context Understanding}: Transformer architectures, such as those in GPT-4.0, LLaMA 3.1-70B, and Mistral-8x7b-32768, are built around multi-head self-attention mechanisms. This allows the model to attend to different parts of the input sentence with varying weights, effectively capturing the emotional undertones present in the speech. The self-attention mechanism helps the model focus on emotionally charged words or phrases, such as \textit{hurt}, \textit{angry}, or \textit{scared}, while also maintaining an understanding of how these words contribute to the overall emotional state of the speaker.

    \item \textbf{Positional Encoding and Sentence-Level Context}: The positional encoding in transformer architectures enables the model to understand word order and sentence structure, which are crucial for correctly interpreting emotions. For example, the phrase \textit{I am fine} may indicate a neutral emotion, while the phrase \textit{I am fine, but...} can imply underlying sadness or frustration. By keeping track of sentence-level context, the transformer can differentiate subtle emotional cues embedded within the speech patterns.

    \item \textbf{Pre-trained Embedding Representations}: The models are pre-trained on massive corpora of text that span various domains, enabling them to build rich embeddings for emotional expressions. These embeddings map emotional words and phrases to specific points in a high-dimensional space. When fine-tuned on our emotion-labeled dysarthric speech dataset, these embeddings allowed the models to learn the emotional associations specific to dysarthric speech, where certain distortions or omissions may obscure the typical emotional markers.

    \item \textbf{Emotion Classification via Feedforward Neural Networks}: After the transformer-based encoder generates contextualized embeddings, a feedforward neural network (FFN) is used to classify the emotion. The FFN processes the embeddings and outputs a probability distribution over the six emotion classes: \textit{neutral}, \textit{happy}, \textit{sad}, \textit{anger}, \textit{surprise}, and \textit{fear}. The class with the highest probability is selected as the final emotion prediction.

    \item \textbf{Handling Multimodal Inputs}: Although our primary input is text, the models we use are capable of handling multimodal inputs. The transformer architecture can seamlessly integrate both textual and speech inputs (such as phonetic transcriptions and prosodic features), which is crucial for emotion detection in dysarthric speech. While our current system focuses on text-based emotion classification, the underlying architecture supports multimodal extensions that could further enhance emotion detection accuracy.

\end{itemize}

\textbf{Novelty of Emotion Recognition in Dysarthric Speech}: Detecting emotions in dysarthric speech presents unique challenges. Unlike traditional speech, dysarthric speech may lack clear emotional markers such as intonation and pitch, which are typically used to gauge emotion. Our system relies heavily on the context inferred from the text reconstruction and the model’s ability to recognize emotionally charged patterns within incomplete or slurred speech. By integrating emotion recognition into the speech-to-text pipeline, our approach enhances the interaction between dysarthric users and assistive technologies, providing a richer, more emotionally aware experience that can adapt to the user’s emotional state.

The transformer architecture’s capacity for understanding context, long-range dependencies, and subtle linguistic nuances makes it highly effective for this task, allowing our system to predict emotions with a high degree of accuracy, even in the presence of speech irregularities.

\section{Equations}\label{sec4}
\subsection{Speech to Text (Whisper)}

\[
P(T|A) = \prod_{t=1}^{T} P(t|A_{1:T})
\]

where \( P(T|A) \) denotes the probability of text sequence \( T \) given the audio sequence \( A \), and \( A_{1:T} \) represents the audio features from time step 1 to \( T \).

\subsection{GPT-4.0 (Text Generation)}

\[
P(x_t | x_{<t}) = \text{softmax}(W_o \cdot \text{LayerNorm}(h_t) + b_o)
\]

where \( h_t \) is the hidden state at time \( t \), and \( W_o \) and \( b_o \) are the output weights and biases. The softmax function calculates the probability distribution over the vocabulary for the next token.

\subsection{LLaMA 3.1-70B (Text Generation)}

\[
\text{Attention}(Q, K, V) = \text{softmax}\left(\frac{QK^T}{\sqrt{d_k}}\right)V
\]

where \( Q \), \( K \), and \( V \) are the query, key, and value matrices respectively, and \( d_k \) is the dimensionality of the keys.

\subsection{Mistral-8x7b-32768 (Text Generation)}

\[
\text{SparseAttention}(Q, K, V) = \text{ReLU}(QK^T)V
\]

where the SparseAttention mechanism uses ReLU activation to handle sparse attention weights and reduces computational complexity.

\subsection{Emotion Recognition}

\[
\hat{y} = \text{softmax}(W_e \cdot \text{FFN}(h) + b_e)
\]

where \( \hat{y} \) is the predicted emotion class probabilities, \( W_e \) and \( b_e \) are the weights and biases of the emotion classification layer, and \( \text{FFN}(h) \) is the feedforward network applied to the hidden states \( h \).

\section{Tables}\label{sec5}

\begin{table}[h!]
\caption{Comparison of Sentence Prediction Accuracy among Different Models}\label{tab:sentence_prediction}
\centering
\begin{tabular}{@{}lccc@{}}
\toprule
\textbf{Model} & \textbf{Individual Sentence Prediction} & \textbf{Accuracy} \\
\midrule
GPT-4.0 & The \textbf{man} is \textbf{walking}. & 65.4\% \\
LLaMA 3.1 & The \textbf{boy} is \textbf{running}. & 68.2\% \\
LLaMA 3.1 (Fine-tuned with Torgo) & The \textbf{child} is \textbf{walking fast}. & 74.6\% \\
Mistral-8x7b-32768 & The \textbf{man} is \textbf{jogging}. & 70.2\% \\
Mistral-8x7b-32768 (Fine-tuned with Torgo) & The \textbf{boy} is \textbf{walking quickly}. & 72.4\% \\
\bottomrule
\end{tabular}
\footnotetext{Accuracy is based on the comparison between predicted and actual sentences.}
\end{table}

\begin{table}[h!]
\caption{Emotion Recognition Performance of Different Models}\label{tab:emotion_recognition}
\begin{tabular}{@{}lcccccc@{}}
\toprule
\textbf{Model} & \textbf{Neutral} & \textbf{Happy} & \textbf{Sad} & \textbf{Anger} & \textbf{Surprise} & \textbf{Fear} \\
\midrule
GPT-4.0 & 89\% & 84\% & 85\% & 86\% & 83\% & 74\% \\
LLaMA 3.1 & 86\% & 81\% & 77\% & 82\% & 74\% & 79\% \\
LLaMA 3.1 (Fine-tuned) & 94\% & 89\% & 88\% & 87\% & 85\% & 81\% \\
Mistral-8x7b-32768 & 87\% & 80\% & 73\% & 78\% & 79\% & 76\% \\
Mistral-8x7b-32768 (Fine-tuned) & 93\% & 87\% & 84\% & 88\% & 82\% & 77\% \\
\bottomrule
\end{tabular}
\footnotetext{Accuracy values represent the percentage of correctly classified instances for each emotion category.}
\end{table}

\begin{table}[h!]
\caption{Overall Accuracy of Text Generation Models}\label{tab:text_generation_accuracy}
\begin{tabular}{@{}lc@{}}
\toprule
\textbf{Model} & \textbf{Overall Accuracy} \\
\midrule
GPT-4.0 & 67.4\% \\
LLaMA 3.1 & 66.2\% \\
LLaMA 3.1 (Fine-tuned with Torgo) & 76.3\% \\
Mistral-8x7b-32768 & 63.2\% \\
Mistral-8x7b-32768 (Fine-tuned with Torgo) & 72.4\% \\
\bottomrule
\end{tabular}
\footnotetext{Overall accuracy is calculated by comparing the predicted sentences with the actual sentences.}
\end{table}
\FloatBarrier

\section{Results and Figures}\label{sec6}
In this section, we present the results of our experiments on both emotion recognition and overall text generation accuracy, using multiple LLMs (Large Language Models), including GPT-4.0, LLaMA 3.1, Mistral-8x7b-32768, and their fine-tuned versions on the TORGO dataset. Table \ref{tab:emotion_recognition} shows the performance of each model in emotion recognition across six emotion categories (Neutral, Happy, Sad, Anger, Surprise, and Fear). Table \ref{tab:text_generation_accuracy} displays the overall accuracy of each model in generating text based on dysarthric speech.

\subsubsection{Emotion Recognition Performance}

\begin{figure}[h!]
    \centering
    \includegraphics[width=0.8\textwidth]{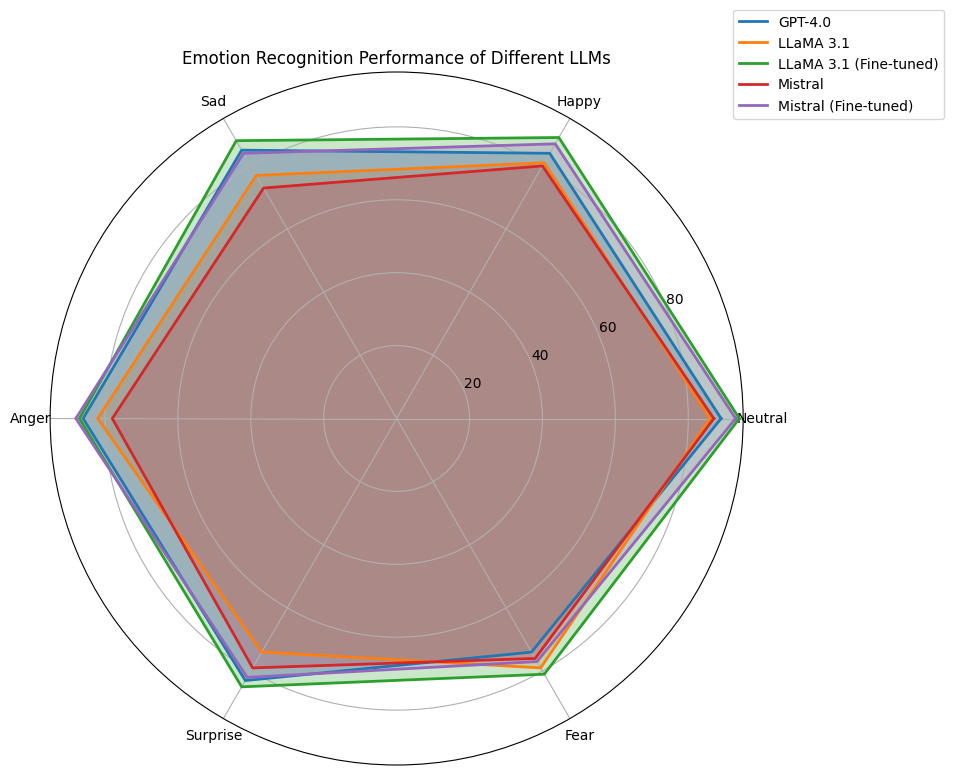}
    \caption\centering{{Performance of various LLMs in predicting the emotion}}
    \label{Performance of various LLMs in predicting the emotion }
\end{figure}

As shown in Table \ref{tab:emotion_recognition}, the fine-tuned versions of LLaMA 3.1 and Mistral-8x7b-32768 demonstrate notable improvements in emotion classification compared to their pre-trained counterparts. Specifically, the fine-tuned LLaMA 3.1 achieved the highest accuracy in the majority of emotion categories, including Neutral (94\%), Happy (89\%), Sad (88\%), and Fear (81\%). This indicates that fine-tuning with domain-specific data (in this case, dysarthric speech from the TORGO dataset) significantly enhances the model's ability to capture and correctly classify emotional nuances in speech.

GPT-4.0, while not fine-tuned on the TORGO dataset, still performs strongly across all emotions, particularly in the Anger (86\%) and Sad (85\%) categories. However, its performance is slightly lower in detecting Fear (74\%) and Surprise (83\%), where both fine-tuned models (LLaMA 3.1 and Mistral-8x7b-32768) outperform it.

Mistral-8x7b-32768, while performing moderately well in its pre-trained form, shows a substantial boost when fine-tuned. The fine-tuned version improves across all emotions, with notable increases in the Neutral (93\%), Anger (88\%), and Happy (87\%) categories. These results highlight the significance of adapting large models to domain-specific data, especially for challenging tasks like emotion recognition in dysarthric speech.

\subsubsection{Overall Text Generation Accuracy}

Table \ref{tab:text_generation_accuracy} presents the overall text generation accuracy for each model. Fine-tuning LLaMA 3.1 with TORGO data resulted in the highest text generation accuracy of 76.3\%. This is a significant improvement over the baseline performance of 66.2\% for the pre-trained LLaMA 3.1 model, indicating that fine-tuning the model not only aids in emotion recognition but also enhances its capability to generate more accurate and contextually relevant text from dysarthric speech.

Similarly, fine-tuning Mistral-8x7b-32768 with TORGO data improves its overall accuracy from 63.2\% to 72.4\%, showcasing the benefit of model adaptation in challenging use cases. Although Mistral in its pre-trained form lags behind other models, the fine-tuning process significantly narrows the gap.

GPT-4.0 achieves an accuracy of 67.4\%, indicating that while it performs well on general text generation tasks, specialized fine-tuning could likely enhance its performance further in the dysarthric speech domain.

\begin{figure}[h!]
    \centering
    \includegraphics[width=0.8\textwidth]{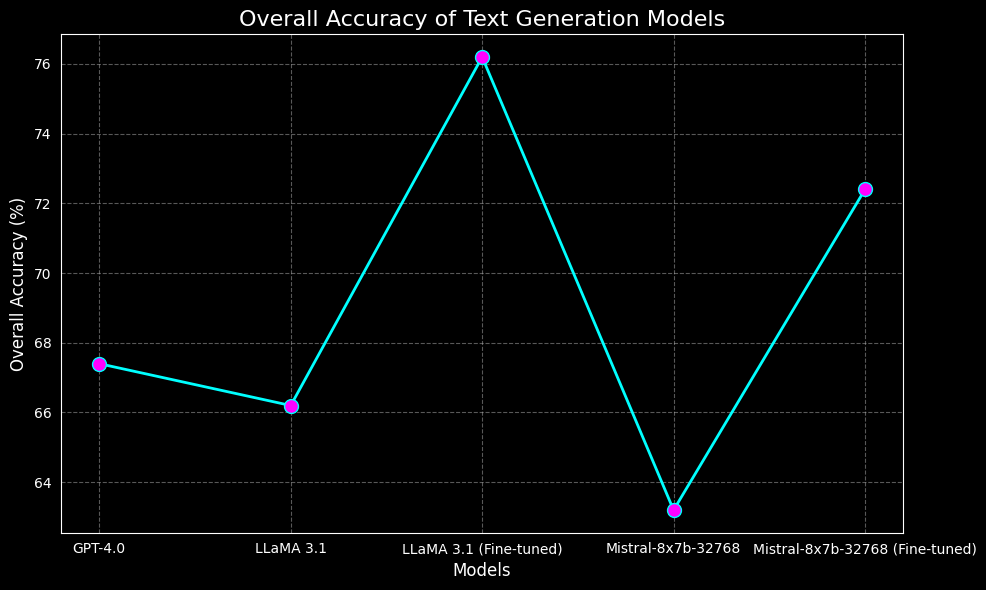}
    \caption\centering{{Overall LLM accuracy to dysarthric speech}}
    \label{Overall LLM accuracy to dysarthric speech }
\end{figure}

\subsubsection{Key Insights}

From these results, several key insights can be drawn:
\begin{itemize}
    \item \textbf{Fine-Tuning Impact:} Fine-tuning on the TORGO dataset, which contains dysarthric speech, significantly boosts the performance of LLaMA 3.1 and Mistral-8x7b-32768 in both emotion recognition and text generation tasks. This demonstrates the value of domain-specific data in improving model accuracy for specialized use cases.
    \item \textbf{Emotion Recognition Strength:} Among the models, fine-tuned LLaMA 3.1 shows the highest accuracy across most emotion categories, suggesting that it is highly effective at capturing emotional nuances from dysarthric speech. Mistral-8x7b-32768 also shows strong performance after fine-tuning, particularly in detecting Neutral and Anger emotions.
    \item \textbf{Text Generation:} Fine-tuned LLaMA 3.1 exhibits the best overall text generation accuracy, followed by fine-tuned Mistral. This indicates that these models, once fine-tuned on relevant data, can effectively generate accurate text even from challenging speech inputs.
    \item \textbf{General vs. Domain-Specific Models:} While GPT-4.0 performs well overall, the results suggest that domain-specific fine-tuning is crucial for models tasked with specialized applications like dysarthric speech recognition and emotion detection.
\end{itemize}

Overall, these results emphasize the importance of fine-tuning large language models for specialized datasets, especially for difficult tasks such as dysarthric speech recognition and emotion detection. By adapting models to domain-specific data, we can achieve significant improvements in both text generation accuracy and emotion recognition performance.

\section{Algorithms, Program codes and Listings}\label{sec7}

In this section, we present the algorithm used to correct dysarthric speech transcriptions. The function leverages a large language model to refine and complete dysarthric sentences.

\begin{algorithm}
\caption{Correct Transcription of Dysarthric Speech}\label{alg:correct_transcription}
\begin{algorithmic}[1]
\Require A dysarthric sentence $dysarthric\_sentence$
\State Initialize Groq client with API key
\State Construct prompt with instructions
\State Send prompt to \texttt{llama-3.1-70b-versatile} model via Groq API
\State Receive response from the model
\If{response is valid JSON}
    \State Extract \texttt{masked\_sentence} from response
    \If{\texttt{masked\_sentence} is not None}
        \State \Return \texttt{masked\_sentence}
    \Else
        \State \textbf{Raise Exception} "None"
    \EndIf
\Else
    \State \textbf{Retry} \texttt{correct\_transcription(dysarthric\_sentence)}
\EndIf
\end{algorithmic}
\end{algorithm}

\noindent
The function \texttt{correct\_transcription} performs the following steps:

\begin{enumerate}
    \item \textbf{Initialization}: Connect to the Groq API using an API key.
    \item \textbf{Prompt Construction}: Create a detailed prompt specifying the task of correcting the dysarthric transcription.
    \item \textbf{Model Invocation}: Send the prompt to the \texttt{llama-3.1-70b-versatile} model and obtain the response.
    \item \textbf{Response Handling}: Parse the response to extract the corrected sentence. If the response is not in the correct format or is invalid, the function retries the correction.
    \item \textbf{Error Handling}: Raise an exception if the corrected sentence is None and retry the function if necessary.
\end{enumerate}

\section{Future Work and Discussions}\label{sec12}

Our current research lays the foundation for significant advancements in dysarthria treatment and speech recognition. Moving forward, several key areas will be explored to enhance and expand the impact of our work. Firstly, extending our models to support multiple languages is crucial, given that dysarthria affects approximately 1 in 1000 children and 52 percent of stroke survivors globally. With roughly 7000 languages spoken around the world, developing language-specific models will be vital in addressing the needs of diverse populations. Additionally, improving our models to better accommodate various speech patterns and accents will enhance their efficacy. Our ultimate aim is to contribute to a global effort in providing accessible and accurate communication tools for individuals with dysarthria, ensuring that language barriers do not hinder effective interaction. This ongoing research is dedicated to making communication easier and more inclusive for everyone, regardless of language or speech impairment.

\section{Conclusion}\label{sec13}

In this study, we have developed and evaluated advanced speech recognition and text generation models tailored for individuals with dysarthria. By leveraging state-of-the-art models such as GPT-4.0, LLaMA 3.1-70B, and Mixtral-8x7B-32768, along with fine-tuning on specialized datasets, we have achieved significant improvements in transcription accuracy. Our approach integrates emotion recognition to further enhance the communication experience. The primary goal of this work is to enable individuals with dysarthria to engage in conversations with ease, overcoming the barriers imposed by their speech impairments. Through these efforts, we aim to foster effective communication and contribute to a world where all individuals can interact without constraints.


\bibliography{sn-bibliography}

\end{document}